\documentclass[review]{elsarticle}

\usepackage{xcolor}
\usepackage{times}
\usepackage{graphicx}
\usepackage{subfigure}
\usepackage{algorithm}
\usepackage{algorithmic}
\usepackage{url}

\usepackage{amssymb}

\usepackage{multirow}
\usepackage{hhline}
\usepackage{threeparttable}

\usepackage{amsmath}
\usepackage{float}
\usepackage{array}

\journal{Computers in Biology and Medicine}

\begin{document}

\begin{frontmatter}

\title{MOMBAT: Heart Rate Monitoring from Face Video using Pulse Modeling and 
Bayesian Tracking}



\author[First]{Puneet Gupta} \ead{puneet@iiti.ac.in}

\author[Second]{Brojeshwar Bhowmick} \ead{b.bhowmick@tcs.com}

\author[Second]{Arpan Pal} \ead{arpan.pal@tcs.com}

\address[First]{Department of Computer Science and Engineering, IIT Indore, Indore, India}
\address[Second]{Embedded system and Robotics, TCS Research and Innovation, Kolkata-700106, India}

\begin{abstract}
A non-invasive yet inexpensive method for heart rate (HR) monitoring is of great importance in many real-world applications including healthcare, psychology understanding, affective computing and biometrics. Face videos are currently utilized for such HR monitoring, but unfortunately this can lead to errors due to the noise introduced by facial expressions, out-of-plane movements, camera parameters (like focus change) and environmental factors. We alleviate these issues by proposing a novel face video based HR monitoring method $MOMBAT$, that is, MOnitoring using Modeling and BAyesian Tracking. We utilize out-of-plane face movements to define a novel quality estimation mechanism. Subsequently, we introduce a Fourier basis based modeling to reconstruct the cardiovascular pulse signal at the locations containing the poor quality, that is, the locations affected by out-of-plane face movements.  Furthermore, we design a Bayesian decision theory based HR tracking mechanism to rectify the spurious HR estimates. Experimental results reveal that our proposed method, $MOMBAT$ outperforms state-of-the-art HR monitoring methods and performs HR monitoring with an average absolute error of 1.329 beats per minute and the Pearson correlation between estimated and actual heart rate is 0.9746. Moreover, it demonstrates that HR monitoring is significantly improved by incorporating the pulse modeling and HR tracking. 
\end{abstract}


\end{frontmatter}


\section{Introduction}
Heart rate (HR) is given by the total number of times a heart contracts or beats per minute. It can assess the human pathological and physiological parameters \cite{nogueira2020analysis}, thus it has attracted the fields of:
i) healthcare;
ii) psychology understanding of stress and mental state;
iii) affective computing for understanding human emotion;
and iv) biometrics for liveness and spoof detection.
These fields can be benefited if HR monitoring is accurate and acquired from an inexpensive sensor in a user-friendly and non-contact manner. This motivates us to propose an accurate HR monitoring method using face videos in this paper.

HR can be measured using contact or non-contact mechanisms. Contact mechanisms require the sensors like electrocardiography (ECG) or 
photo-plethysmography (PPG). They can mitigate illumination artifacts and provide synchronized multi-modal physiological data, but they should be properly placed on the body. In real-world scenarios, motion can change the
contact area between the sensor and human skin, which eventually results in spurious HR monitoring \cite{zhang2015troika}. Since these methods require the contact between the user and the sensor for large duration, they restrict unobtrusive monitoring and require a dedicated sensor for single user monitoring. Also, maintaining the sensor contact is cumbersome for: i) neonates surveillance;
ii) analyzing sleep quality;
iii) exercise monitoring during rehabilitation etc. \cite{Broj1} \cite{Broj3} ; and
iv) observing skin damaged patients.
These issues can be handled by performing HR monitoring using non-contact mechanisms, which allow the monitoring anytime and anywhere with minimal user involvement. These non-contact mechanisms can also be used for
covert monitoring and thereby utilized for sleep monitoring,  lie detection \cite{owayjan2012design} and stress monitoring \cite{salahuddin2007ultra}.
Due to these advantages, non-contact based HR monitoring is proliferating.

Traditional non-contact mechanisms require bulky, expensive and dedicated sensors like Microwave Doppler and laser for HR monitoring \cite{huang2016self}. Modern non-contact mechanisms employ inexpensive and
portable camera sensors for HR estimation. They are based on the phenomenon that heart beats generate the cardiovascular pulse which propagates in the entire human body. It introduces color variations in the reflected light \cite{gupta2017accurate} and micro-movements in the face \cite{balakrishnan2013detecting}. Both these contain the cardiovascular pulse information and are imperceptible to the human eye, but they can be analyzed using the camera for estimating the HR.

Existing face based HR methods analyze the micro-motion or color variations across time and refer to them as temporal signals \cite{gupta2017accurate}. The cardiovascular pulse is estimated from the temporal signals and it is eventually used for HR estimation. Along with the subtle pulse signal, the temporal signal 
constitutes prominent noise originated from: i) facial expression;
ii) eye blinking;
iii) face movements; 
iv) respiration;
v) camera parameters (for example, change in focus); and
vi) environmental factors (for example, illumination variations). Extraction of HR signal from such a noisy temporal signal is thus a challenging problem. In this paper, we alleviate these issues to improve the face videos based HR monitoring by introducing a novel method $MOMBAT$, that is, MOnitoring using Modeling and BAyesian Tracking. The main research contributions of our proposed method, $MOMBAT$ are:i) it introduces a novel quality estimation mechanism that adapts according to
the out-of-plane face movements and provide quality of each frame, unlike
existing mechanisms that provide single quality for the entire video; ii) it initiates the utilization of Fourier basis based pulse modeling for reconstructing the pulse signals at the poor quality video frames using the pulse signals at the good quality video frames; and iii) it presents a novel Bayesian decision framework for rectifying the spurious HR estimates. 

The paper is organized as follows.The preliminaries required for better understanding of our method, $MOMBAT$are discussed in Section \ref{prelim} and $MOMBAT$ is presented in Section \ref{proposedMethod}. The experimental results are analyzed in Section \ref{Experimental_Results} followed by conclusions in the last section.

\section{Preliminaries}   \label{prelim}

\subsection{Face HR Estimation}
Typically, any face videos based HR estimation method consists of the following three stages; preprocessing, HR estimation, and post-processing.

\subsubsection{Preprocessing} 
During preprocessing, a region of interest (ROI) containing useful pulse information is detected. Skin pixels contain pulse information, thus face detection followed by removing non-skin pixels are performed for ROI extraction \cite{rapczynski2018region}. Usually, a face is detected using Viola-Jones \cite{viola2001rapid} or model based \cite{baltrusaitis2013constrained} face detectors. Subsequently, non-skin pixels due to background and hairs, are removed by applying skin color discrimination techniques. Inevitable movements (like eye blinking) near the eye areas can degrade the HR estimation \cite{gupta2017serial}.  Thus, the eye areas are detected by employing facial geometry heuristics or trained classifiers \cite{wang2005automatic} and then these eye areas are removed for the better estimation. The remaining face area is used to define the region of interest (ROI). Some commonly used ROI are full face, forehead region  or cheek areas \cite{rapczynski2018region}. ROI locations can be shifted by the  facial movements in z-direction known as out-of-plane transformations or movements in x and y-dimensions known as in-plane transformations. Both these transformations can result in spurious HR estimation due to ROI shifting. Hence, these transformations are minimized using face registration for improving the HR estimation \cite{gupta2018robust}. One can use mobile  based 3D depth estimation also to get the depth of landmark \cite{Broj2} \cite{Broj4} for compensating out-of-plane movements. We use simple distance between the eyes is used by \cite{gupta2018robust} for the registration, but it can be spurious due to eye-blinking. These transformations can be accurately measured  by wearables \cite{borghi2018face}, but it requires human contact and thus, avoided for non-contact face video based HR.

\subsubsection{Temporal Signal Extraction} 
Micro-motion and subtle color variations in the  face video can be determined using Lagrangian \cite{balakrishnan2013detecting} and  Eulerian techniques \cite{poh2011advancements} respectively.These variations across different frames provide temporal signals.In Lagrangian techniques, discriminating features are extracted from the ROI and they are explicitly tracked in the subsequent frames for determining the temporal signals \cite{tang2019removing}. This tracking is not only time-consuming, but also spurious due to improper illumination. Alternatively, temporal signals can be determined using Eulerian techniques, where color variations are examined in the fixed ROI \cite{poh2011advancements}. The Eulerian techniques are less time-consuming than the Lagrangian techniques, but they are applicable only when small variations are present \cite{Wu12Eulerian}. It requires fixed ROI and hence, altered tremendously even if the face is slightly moved \cite{gupta2018exploring}. 

Eulerian temporal signals are given by the color variations in the face video, having RGB color channels. Amongst these channels, the green channel contains the strongest photo-plethysmographic signal because: i) haemoglobin absorbs green light better than red,  which makes green light less susceptible to motion noise as compared to red light; and ii) green light penetrates sufficiently deeper into the skin as compared to blue light \cite{verkruysse2008remote}. It is apparent that better performance can be expected by fusing all RGB color channels. Model based methods utilize optical and physiological properties of skin reflection to perform such a fusion. Unfortunately, such methods are not applicable in all possible scenarios. For example, well known model based methods, CHROM \cite{de2013robust} and POS \cite{wang2016algorithmic} do not provide correct HR estimation when pulse signal and noise share similar amplitudes \cite{wang2016algorithmic}. Furthermore, POS fails when face videos are acquired under inhomogeneous illumination conditions, that is, when faces are illuminated by  multiple light sources \cite{wang2016algorithmic}.

\subsubsection{HR Estimation} 

\begin{figure}	
	{
		\includegraphics[scale = 0.7]{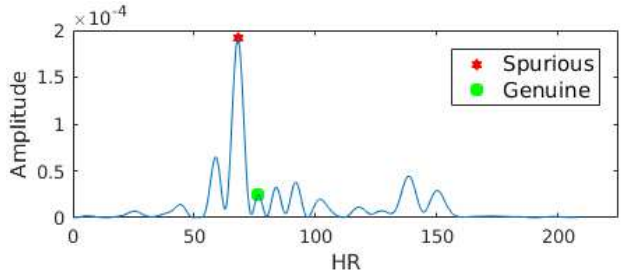}
	}	
  \caption{Spectrum of the temporal signal containing noise. 
HR (in bpm) and their corresponding amplitude 
are depicted using X and Y-axes respectively.}
\label{fig:noise_effect}
\end{figure}

Pulse signal is estimated from the temporal signal using statistical learning. As an instance, periodicity analysis and blind source separation (BSS) techniques are used for the pulse signal estimation by \cite{mcleod2017analysis} and \cite{gupta2018robust} respectively. Usually, Fast Fourier Transform (FFT) is applied to the pulse signal and the frequency corresponding to the maximum amplitude in the pulse spectrum corresponds to the HR \cite{camm1996heart}. But when the temporal signal is contaminated with noise, several spurious peaks are generated and the actual HR may not correspond to the maximum amplitude peak. An example is shown in Figure \ref{fig:noise_effect} where several spurious peaks are generated due to facial movements. Several filtering techniques can be employed to remove the noise in the temporal signals and thereby improve HR estimation. For example, Detrending filter is applied to alleviate the non-stationary trend in the pulse signal \cite{tarvainen2002advanced}. Some spectrum subtraction methods that mitigate the noise from the pulse signal are proposed by \cite{huang2016accurate, li2014remote}; and \cite{lee2015heart}. The noise due to motion artifact is estimated by \cite{huang2016accurate} using facial boundary tracking. Such tracking is spurious due to facial pose variations. Similarly, background variations and brightness are estimated by \cite{li2014remote} and \cite{lee2015heart} respectively, for estimating the noise due to illumination variations. But they are highly dependent on the background characteristics and the distance between the background and face \cite{lam2015robust}.

\subsection{HR Monitoring}
HR monitoring continuously performs HR estimations at different small time intervals and, usually, facial deformations affect a small number of frames. Due to this, some HR estimates in the monitoring can be spurious due to the inevitable facial deformations. Better estimation can be expected when a large number of frames are considered \cite{tulyakov2016self}. But it results in the loss of HR variations which is highly useful for medical purposes \cite{berntson1997heart}. Furthermore, it restricts user-friendliness due to high wait time. Typically, the number of frames in a time interval is chosen such that the cardiovascular pulse wave can complete at least two cycles.

HR monitoring is performed by \cite{rodriguez2018video} using green channel variations and band-pass filtering. Likewise, methods \cite{tulyakov2016self} and \cite{qiu2018evm} perform HR monitoring using matrix completion and convolution neural network (CNN) respectively. Erroneous HR estimates are rectified by \cite{gupta2018robust} to improve the HR monitoring using image registration and global HR. The global HR is estimated from all the video frames and thus, it can be spurious when temporal signals contain noise.

\subsection{Constrained Local Neural Field (CLNF)}
Detecting the discriminatory facial features is an extensively studied research topic. These are referred to as facial landmark points \cite{wang2014facial}. Usually, they are located around face boundaries, eyes, eyebrows, mouth and nose. Constrained Local Model (CLM) is highly useful for landmark detection. It consists of: i) point distribution model (PDM) that uses rigid and non-rigid shape transformations for modelling the global location of discriminatory points; ii) patch experts which models the behaviour of a landmark by analysing the appearance around its local neighbourhood; and iii) joint optimization which aims to fit PDM and the experts in the best possible way \cite{saragih2011deformable}. The unknown shape parameter $\boldsymbol{p}$ is estimated by the joint optimization, which is given by
\begin{equation}
\boldsymbol{p}^* =  \arg\min_{\boldsymbol{p}}\left [ R\left ( \boldsymbol{p} \right ) + \sum_{i=1}^{n}D_i\left ( \mathrm{\boldsymbol{x}_i},I \right ) \right ]
\end{equation}
where $R$ is the regularization term which restricts the introduction of unlikely shapes; $D_i$ is the misalignment in the location of $i^{th}$ landmark in the image $I$; and $\mathrm{\boldsymbol{x}_i}$ is the $i^{th}$ landmark location in 3-D which is given by 
\begin{equation}
\mathrm{\boldsymbol{x}_i} = \mathit{\boldsymbol{s}} \cdot \boldsymbol{R} \cdot \left ( \mathrm{\bar{\boldsymbol{x}_i}} + \phi_i \boldsymbol{q} \right )  + \boldsymbol{t}
\end{equation}
where $\mathrm{\bar{\boldsymbol{x}_i}}$ denotes the mean value of $i^{th}$ feature given by PDM; $\phi_i$ is the component matrix; and vector $\boldsymbol{q}$ is used to control the non-rigid shape \cite{baltruvsaitis20123d}. Remaining parameters scaling $\mathit{\boldsymbol{s}}$, translation $\boldsymbol{t}$ and rotation $\boldsymbol{R}$ controls the rigid shape. In essence, shape parameters are given by $\boldsymbol{p} = \left [ \mathit{\boldsymbol{s}}, \boldsymbol{t}, \boldsymbol{R}, \boldsymbol{q} \right ]$. The performance of CLM heavily relies on PDM, patch expert and joint optimization. In CLNF \cite{baltrusaitis2013constrained}, patch experts are given by local neural field for modeling spatial relationships between pixels, while non-uniform regularized landmark mean-shift is proposed for joint optimization by taking into account the reliability of patch experts. 

\subsection{Pulse Extraction using Kurtosis Optimization}

Each temporal signal contains a pulse signal along with the noise. In case of multiple temporal signals, the pulse signal is extracted using blind source separation by estimating the individual source components \cite{poh2011advancements}. Amplitudes of pulse signal and noise in the temporal signals depend on the facial structure, user characteristics (like skin color) and environmental settings (like illumination). Hence, z-score normalization \cite{ross2006handbook} is applied to normalize the temporal signals. Moreover, the temporal signal, $F^i$ contains noise, $\boldsymbol{\eta}$ and actual pulse signal, $\boldsymbol{X}_a$ but modified by the facial structure. That is,
\begin{equation}
\label{basic}
F^i\left ( n \right ) = A\boldsymbol{X}_a\left ( n \right ) + \boldsymbol{\eta}\left ( n \right )
\end{equation}
where $n$ and $A$ denote the time instant and matrix incorporating the effects of facial structure respectively. Further, the actual pulse signal is not known and it requires estimation from the temporal signal, that is,
\begin{equation}
\label{basic1}
\boldsymbol{X}_e\left ( n \right ) = BF^i\left ( n \right )
\end{equation}
where $\boldsymbol{X}_e$ and $B$ denote the estimated actual pulse and the transformation matrix respectively. It can be observed from Equations \eqref{basic} and \eqref{basic1} that:
\begin{equation}
\label{finalpulserelation}
\boldsymbol{X}_e\left ( n \right ) = T\boldsymbol{X}_a\left ( n \right ) + \hat{\boldsymbol{\eta}}\left ( n \right )
\end{equation}
such that $T$ = $BA$ and $\hat{\boldsymbol{\eta}}$ = $B{\boldsymbol{\eta}}$.

For accurate HR monitoring, $\boldsymbol{X}_e$ should be similar to $\boldsymbol{X}_a$. That is, magnitude of $T$ should be 1 and appropriate shape constraints should be imposed on the estimated pulse spectrum. Such shape constraints are imposed using higher order cumulants \cite{reichert1992automatic}. The highest order of cumulant is restricted to 4 because higher-order cumulants are easily affected by the tail of the distribution which makes them sensitive to outliers  and they are slightly independent in the middle of the distribution containing useful information \cite{huber1985projection}. It is proved in \cite{papadias2000globally} that constraints on cumulant similarities till fourth order can be fulfilled by defining the objective function as:
\begin{equation}
\label{obj_fun}
\max_{T} \ \ \left |K\left [\boldsymbol{X}_e  \right ]  \right | \ \ subject\  to \ \ T^*T=1
\end{equation}
where $K\left [\bullet  \right ]$ denotes the Kurtosis \cite{decarlo1997meaning} while $\left | \bullet   \right |$ and $^*$ represent the absolute value and conjugate operations respectively. This Kurtosis based maximization is solved using \cite{papadias2000globally} to obtain the estimated pulse signal because it quickly provides the global convergence.

\subsection{Quality estimation of pulse signal}

Quality of pulse signal can be estimated using the peak signal to noise ratio (PSNR) \cite{yu2006method}. Typically, the amplitude of the pulse spectrum obtained after converting the pulse signal into the frequency domain, should  contain a peak at the HR frequency and negligible values at other frequencies. Unfortunately, in the pulse spectrum, the noise increases the amplitude at other frequencies. Thus, PSNR can be defined such that the signal can be interpreted  as the amplitudes corresponding to HR frequency while noise can be thought of the amplitudes at the remaining frequencies. Mathematically, the quality given by PSNR, $q$ is given by:
\begin{equation}
\label{eq:PSNR}
q = \frac{\sum_{i = maxLoc(\boldsymbol{S}_p)-n_p}^{maxLoc(\boldsymbol{S}_p)+n_p}\boldsymbol{S}_p\left ( i \right )}{\bar{sum}(\boldsymbol{S}_p) - \sum_{i = maxLoc(\boldsymbol{S}_p)-n_p}^{maxLoc(\boldsymbol{S}_p)+n_p}\boldsymbol{S}_p\left ( i \right )}
\end{equation}
where $\boldsymbol{S}_p$ denotes the spectrum of the estimated pulse signal; $\bar{sum}$ performs the sum over all the frequencies;  $n_p$ represent the neighbourhood size; and $maxLoc$ returns the position containing the maximum value (thus, the location of HR frequency is given by $maxLoc(\boldsymbol{S}_p)$). In equation \eqref{eq:PSNR}, signal (or numerator) is obtained by adding the amplitude of HR frequency and its few neighbourhoods while noise (or denominator) is obtained by adding the amplitude of the remaining frequencies.

\begin{figure}	
	{
		\includegraphics[scale = 0.35]{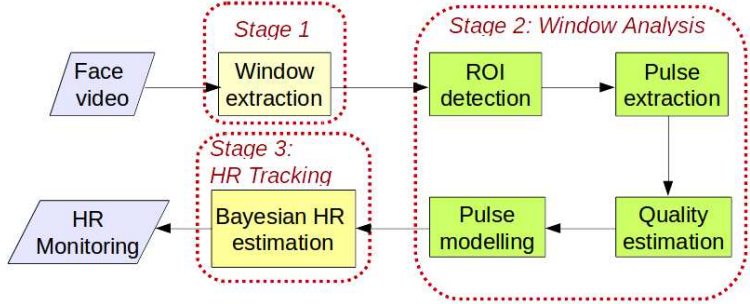}
	}	
  \caption{Flow-graph of our Proposed Method, $MOMBAT$}
  \label{fig:Flow}
\end{figure}

\section{Proposed Method}           \label{proposedMethod}

Our proposed face based HR monitoring method, $MOMBAT$ is presented in this section. It consists of the following three stages: window extraction, window analysis and HR tracking. In the first stage, we divide the face video into several overlapping windows. In the next stage, we estimate the cardiovascular pulse and quality for each window. Subsequently, we introduce pulse signal modeling to obtain better HR estimates. In the last stage, we propose Bayesian tracking to improve the HR monitoring.  Figure \ref{fig:Flow} illustrates the flow-graph of the proposed method, $MOMBAT$.

\subsection{Window Extraction}

HR monitoring requires the estimation of multiple HR at various time intervals and eventually concatenation of all HR estimates. Hence, just like the existing HR monitoring methods, we divide the face video into multiple overlapping windows \cite{gupta2018robust}.

\subsection{Window Analysis}   

In this section, we analyze each extracted window to estimate the corresponding cardiovascular pulse, HR and quality. Initially, we detect ROIs from the window and mitigate in-plane face movements. Then, we extract the cardiovascular pulse from the ROIs using Eulerian technique followed by FFT based analysis. Subsequently, we estimate the quality of the pulse according to their out-of-plane deformations and utilize it to rectify the pulse using pulse modeling.

\begin{figure}	
		{
		\includegraphics[scale = 0.25]{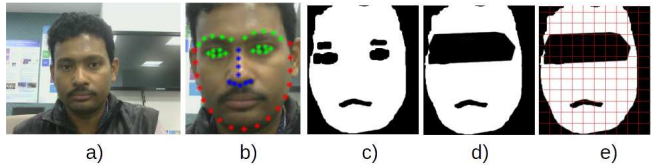}
	}
  \caption{Steps in ROI Detection: a) Video frame;  
  b) Landmarks on the face using CLNF; c) Detected skin mask;
  d) Resultant skin mask; and
  e) Selected ROIs from resultant mask.
  (Figures best viewed in colors)}
  \label{fig:ROI_Detection_Steps}
\end{figure}

\subsubsection{ROI Detection}    \label{ROI_Detection}
The facial skin area contains useful pulse information, hence we utilize it to define ROI. Initially, we detect the facial areas and landmarks using Constrained Local Neural Field (CLNF) model proposed by \cite{baltruvsaitis2016openface}. HR can be spurious when non-skin pixels (like beard) and eye areas are utilized for HR estimation \cite{gupta2017accurate}. Thus, we detect these areas and remove them. We utilize skin detection proposed by \cite{phung2002novel} to detect the non-skin pixels and we obtain the eye area by the convex hull of facial landmarks corresponding to the eyes and eyebrows. Furthermore, subtle motion in facial boundaries can significantly alter the temporal signals and thereby result in spurious HR. Hence, we remove the boundary pixels by performing morphological erosion \cite{gupta2015accurate}. Figure \ref{fig:ROI_Detection_Steps} illustrates these steps.

\begin{figure}	
		{
		\includegraphics[scale = 0.3]{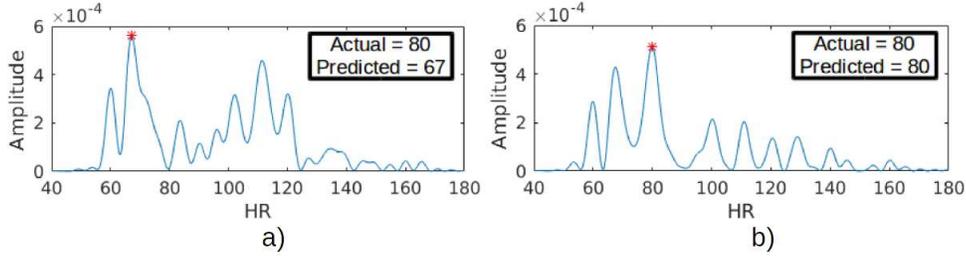}
	}
  \caption{Usefulness of Image Registration: a) Pulse obtained without   frame registration; and b) Pulse obtained after image registration. HR (in bpm) and their corresponding amplitude  are depicted using X and Y-axes respectively.}  
\label{fig:Frame_reg_compare}\end{figure}

The translation and rotation of face in x and y-dimensions, known as in-plane transformations, can shift the location of the ROI in subsequent frames and thereby alter the Eulerian temporal signals and results in the spurious HR estimation. It motivates us to perform image registration between subsequent frames, so as to mitigate the in-plane transformations. An example depicting the applicability of image registration is shown in Figure \ref{fig:Frame_reg_compare}. It shows the pulse spectrum before and after applying the image registration in Figures \ref{fig:Frame_reg_compare}(a) and \ref{fig:Frame_reg_compare}(b) respectively. It can be observed that HR can be correctly estimated after employing image registration. We perform the registration between subsequent frames by minimizing the deviation between nose landmark points because nose area is least affected by the facial expressions. Figure \ref{fig:ROI_Detection_Steps}(b) shows the chosen landmark points in blue color. Mathematically, we first estimate the transformation matrix, $\bar T$ between the current and previous frames using:
\begin{equation}
\label{opt_func}
  \bar T =  \arg\min_{T}\left [ \sum_{i=1}^{q}  \left \| \boldsymbol{F}_i-T \left ( \boldsymbol{M}_i \right )  \right \| _2\right ]
  \end{equation}
where $\boldsymbol{M}_i$ and $\boldsymbol{F}_i$ denote the positions of $i^{th}$ nose landmark point in current and previous video frames respectively; $q$ is the total number of nose landmark points; and $T$ is the transformation matrix consisting of translations and rotation in 2-D, that is:
\begin{equation}
 T = 
\begin{bmatrix}
cos\left ( \theta \right ) & -sin\left ( \theta \right ) & t_x \\ 
 sin\left ( \theta \right )& cos\left ( \theta \right ) & t_y\\ 
 0&0  & 1
\end{bmatrix}
\end{equation}
where $\theta$ is the rotation angle while $t_x$ and $t_y$ are the translations in x and y directions respectively. It is important to note that $\boldsymbol{M}_i$ and $\boldsymbol{F}_i$ denotes the feature points positions in homogeneous coordinates, that is, the feature at $\left [ x,y \right ]^T$ is represented by $\left [ x,y,1 \right ]^T$. We utilize Gradient Descent optimization  to solve the Equation \eqref{opt_func} \cite{gupta2015accurate}. The in-plane transformation is minimized by registering the current image using:
\begin{equation}
 R = \bar T \left ( I_M \right )
\end{equation}
where $R$ is the registered image and $I_M$ is the current video frame. Thereafter, facial expressions can also result in spurious HR estimation. It can  be mitigated by considering several face areas as different ROIs rather than considering full face as one ROI \cite{gupta2017accurate}. Hence, we utilize the method proposed by \cite{gupta2017accurate} for ROI extraction. For brevity, it divides the resultant registered face area into non-overlapping square blocks and considers them as ROIs. Also, it chooses the block-size such that  the  detected  area should contain 10  blocks in the horizontal  direction. An example is shown in Figure \ref{fig:ROI_Detection_Steps}(e). 

\subsubsection{Pulse Extraction}     \label{pulse_Signal}

We estimate the cardiovascular pulse using the method proposed by \cite{gupta2017accurate}. For brevity, it first extracts the Eulerian temporal signals from each ROI using the variations introduced in the average green channel intensities because the green channel contains the strongest plethysmographic signal amongst RGB color channels. Mathematically, the temporal signal $\boldsymbol{S}^i$ corresponding to $i^{th}$ ROI is given by:
\begin{equation}
\boldsymbol{S}^i = \left [ s^i_1, s^i_2, \cdot \cdot \cdot \cdot s^i_{\left (f-1 \right )}\right ]
\end{equation}
where $f$ is the total number of frames and $s^i_k$ representing the variations in $k^{th}$ frame for $i^{th}$ ROI is given by:
\begin{equation}
s^i_k = \sum_{\left ( x,y \right )\in B^i}\left ( F_{\left (k+1 \right )}^g\left ( x,y \right )  - F_k^g\left ( x,y \right ) \right )
\end{equation}
where $B^i$ represents the $i^{th}$ ROI; $\left ( x,y \right )$ denotes a pixel location; and $F_k^g$ stores green channel intensities in $k^{th}$ frame.
The extracted temporal signals contain noise which is mitigated by utilizing a band-pass filter and a Detrending filter \cite{gupta2017accurate}. The cardiovascular pulse,$\bar{\boldsymbol{X}}_e$ is eventually extracted by applying  the kurtosis based optimization proposed in \cite{gupta2017accurate}.

\subsubsection{Quality estimation}

\begin{figure*}	
\centering
		{
		\includegraphics[scale = 0.45]{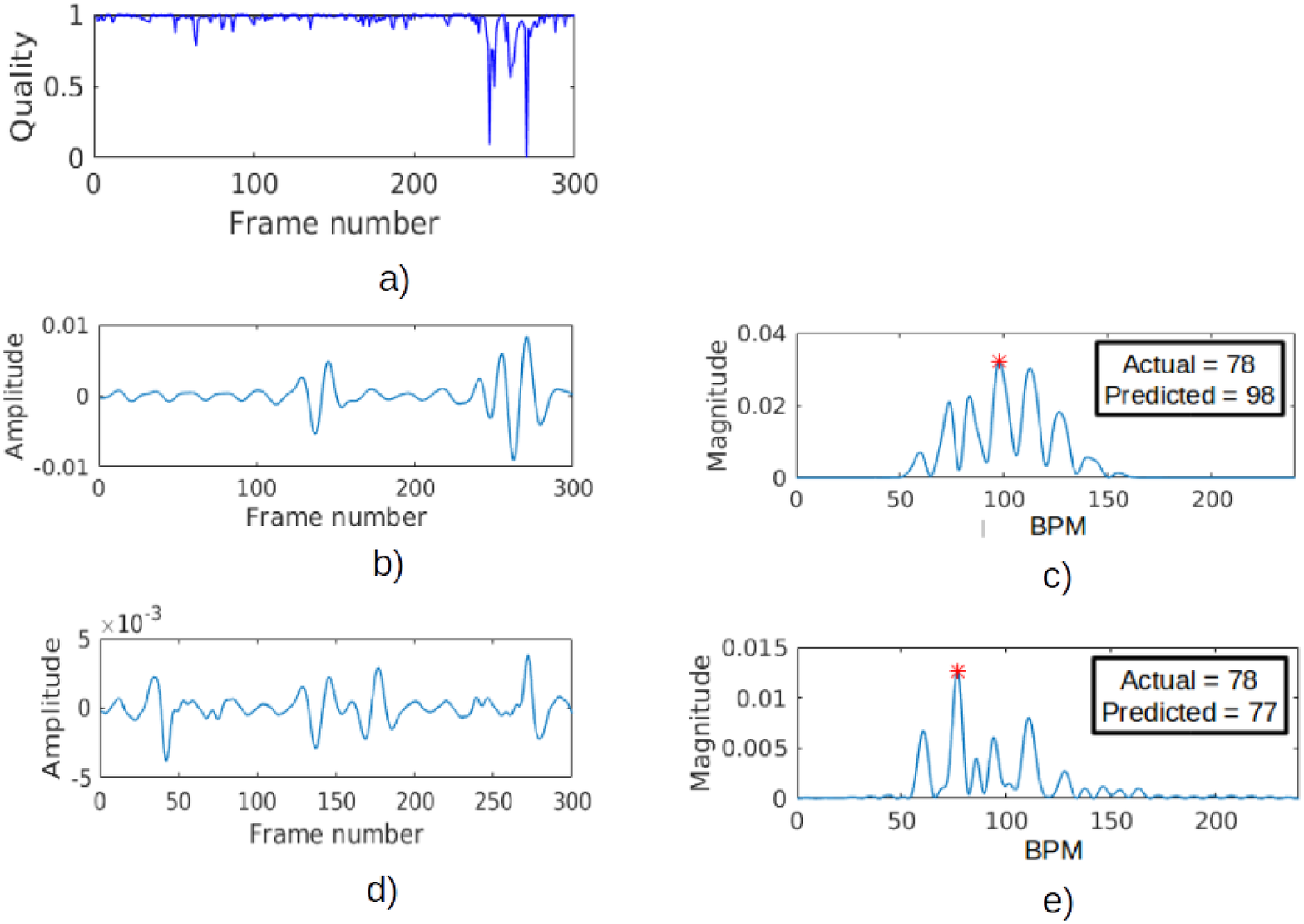}
	}
  \caption{Steps in post-processing: a) Quality, $\hat Q$; pulse signal, $\bar X_e$ and its spectrum without post-processing are shown in b) and c) respectively; while pulse signal, $\hat X_e$ and its spectrum after post-processing are shown in d) and e) respectively. For illustrating the spectrum in c) and e), we depict the HR (in bpm) and their corresponding amplitude using X and Y-axes respectively.}
  \label{fig:Post_processing_steps}
\end{figure*}

Just like in-plane deformations, out-of-plane deformations caused by facial movements in z-direction, can shift the ROI and result in the spurious HR estimation. We introduce a novel quality measure which incorporates these out-of-plane movements to measure the confidence in the correct estimation of pulse signal at each frame. It is defined using the 3-D facial landmarks that we have detected by applying Constrained Local Neural Field (CLNF) model \cite{baltruvsaitis2016openface} in Section \ref{ROI_Detection}. Amongst these, we utilize only the 3-D facial landmark points corresponding to the face boundary for detecting the out-of-plane movements because the face boundary is highly affected by the motion in the z-direction. These selected landmark points are shown in red color in Figure \ref{fig:ROI_Detection_Steps}(b). Out of these chosen points, the points containing the largest deviation in z-direction are used for the quality estimation. It can be observed that yaw head motion can move some boundary points in positive and some in negative z-directions, thus we evaluate the deviation using maximum absolute change in z-direction. In essence, the deviation in the $k^{th}$ frame, $d_{\left ( k-1 \right )}$ is given by:
\begin{equation}
 d_{\left ( k-1 \right )} = max \left (\left | l^j_k - l^j_{\left (k-1 \right )} \right | \right )\end{equation}
where $max$ is the maximum operator; $\left | \bullet \right |$ is the absolute operator; while $l^j_k$ and $l^j_{\left (k-1 \right )}$ denote the z-coordinate of $j^{th}$ landmark in $k^{th}$ and $\left (k-1 \right )^{th}$ frames respectively. After evaluating the deviations for all the frames, except the last frame, we compute the quality at $\left ( k-1 \right )^{th}$ frame, $\hat q_{\left ( k-1 \right )}$ using:
\begin{equation}
\label{eq:q_def}
\hat q_{\left ( k-1 \right )} = 1 - \left (\frac{d_{\left ( k-1 \right )} - min\left ( \boldsymbol{d} \right )}{max\left ( \boldsymbol{d} \right ) - min\left ( \boldsymbol{d} \right )}  \right )
\end{equation}
where $min$ is the minimum operator and $\boldsymbol{d}$ stores all the computed deviations, that is
\begin{equation}
 \boldsymbol{d} = \left [ d_1, d_2, \cdot \cdot \cdot, d_{f-1} \right ]
\end{equation}
where $f$ is the number of frames. In essence, $d_{\left ( k-1 \right )}$ in Equation \eqref{eq:q_def} is first normalized to $\left [  0,1\right ]$ and then modified to define quality such that low and high deviations corresponds to high and low quality values respectively. Thus, the quality due to out-of-plane movements, $\hat{\boldsymbol{Q}}$ is given by:
\begin{equation}
\label{quality_1}
 \hat{\boldsymbol{Q}} = \left [ \hat q_1, \hat q_2, \cdot \cdot \cdot , \hat q_{\left ( f-1 \right )}   \right ]
\end{equation}
An example of the quality estimation using out-of-plane movements is shown in Figure \ref{fig:Post_processing_steps}(a).

\subsubsection{Pulse Modeling}

The frames affected by out-of-plane movements provide spurious temporal signals and thereby results in an incorrect estimation of cardiovascular pulse, $\bar X_e$. An example of such pulse is shown in Figure \ref{fig:Post_processing_steps}(b) along with its corresponding pulse spectrum in Figure \ref{fig:Post_processing_steps}(c). It can be observed from Figure \ref{fig:Post_processing_steps}(c) that the predicted HR is deviated significantly from the actual HR. To improve the efficacy of the pulse signal, we introduce Fourier basis based modeling that aims to reconstruct the pulse signal at those frames which are affected by out-of-plane movements. We formulate the problem of noise reduction as a data fitting problem \cite{gupta2016accurate}. It consists of the following steps: i) defining appropriate basis functions; ii) parameter fitting; and iii) signal reconstruction. Mathematically, $\bar{\boldsymbol{X}}_e$ can be decomposed as:
\begin{equation}
\label{model_1}
 \bar{\boldsymbol{X}}_e\left ( x \right ) \approx \sum_{i=1}^{\alpha} a_i \boldsymbol{\phi}_i\left ( x \right )
\end{equation}
where $\alpha$ is the number of basis; $a_i$ denotes the model parameter for $i^{th}$ parameter; $x$ denote the frame number; and $\boldsymbol{\phi}_i\left ( x \right )$ is the $i^{th}$ basis function. Parameter $\alpha$ plays a crucial role in the modeling. Pulse reconstruction is spurious when $\alpha$ is small and if it is set to high value, then even noise can be modeled. We describe the parameter selection of $\alpha$ in Section \ref{ParameterSelection}. For simplicity, Equation \eqref{model_1} can be written in matrix form using:
\begin{equation}
\label{model_2}
 \bar{\boldsymbol{X}}_e \approx \boldsymbol{A} \Phi
\end{equation}
where
\begin{equation}
 \Phi = 
\begin{bmatrix}
\boldsymbol{\phi}_1\left ( 1 \right )& \boldsymbol{\phi}_2\left ( 2 \right ) & \cdot & \cdot \ & \boldsymbol{\phi}_1\left ( f-1 \right )  \\ 
\boldsymbol{\phi}_2\left ( 1 \right )& \boldsymbol{\phi}_2\left ( 2 \right ) & \cdot & \cdot \ & \boldsymbol{\phi}_2\left ( f-1 \right )  \\ 
\cdot & \cdot & \cdot \ & \cdot  & \cdot \\ 
\cdot & \cdot & \cdot \ & \cdot  & \cdot \\ 
\boldsymbol{\phi}_{\alpha}\left ( 1 \right )& \boldsymbol{\phi}_{\alpha}\left ( 2 \right ) & \cdot & \cdot \ & \boldsymbol{\phi}_{\alpha}\left ( f-1 \right ) 
\end{bmatrix}
\end{equation}
and
\begin{equation}
 \boldsymbol{A} = \left [ a_1 \   a_2 \ \cdot \cdot  \cdot \ a_{\alpha}\right ]
\end{equation}
We define basis functions, $\Phi$, using the well known Fourier basis \cite{gupta2016accurate}. It is because i) these basis are orthogonal which is required to provide stability in the optimization by assuring low residual error; and ii) their amplitude lies in the range of $\left [ -1,1 \right ]$ which helps in avoiding the problem of overflowing integer with polynomial basis. The Fourier basis for the order $n$ are given by:
\begin{equation}
\boldsymbol{\phi}_{\left ( n \right )}\left ( x \right ) = \left\{\begin{matrix}
sin\left ( \frac{\left (n+1  \right )}{2}\times x \right ) \ \ \mbox{if } n \mbox{ is odd}
\\ cos\left ( \frac{n}{2}\times x \right ) \ \ \mbox{if } n \mbox{ is even}
\end{matrix}\right.
\end{equation}
This represents an overdetermined system of linear equations because the small number of unknown parameters, $\alpha$ needs to be estimated from a large number of observations, $\left ( f-1 \right )$. Furthermore, we aim to reconstruct the pulse at the frames  containing large out-of-plane movements by utilizing the pulse information at the frames containing small out-of-plane movements. Thus, we solve this overdetermined system of linear equations using weighted least square estimation where weights are given by  quality due to out-of-plane movements, $\hat{\boldsymbol{Q}}$ \cite{gupta2016accurate}. That is,
\begin{equation}
\label{WLS}
\mathbf{\hat{A}} = \arg\min_{\boldsymbol{A}}\left \| \hat{\boldsymbol{Q}}\left (\bar{\boldsymbol{X}_e} - \boldsymbol{A} \Phi  \right ) \right \|_2
\end{equation}
where $\mathbf{\hat{A}}$ contains the estimated modeling parameters; $\left \| \bullet \right \|$ contains the norm; and $\hat{\boldsymbol{Q}}$ (solved in Equation \eqref{quality_1}) is the quality due to out-of-plane movements. The solution of Equation \eqref{WLS} is given by:
\begin{equation}
 \mathbf{\hat{A}} = \left ( \Phi^T \bar Q \Phi  \right )^{-1}\Phi^T \bar Q \bar{\boldsymbol{X}_e}
\end{equation}
where $\bar Q$ is the diagonal matrix formed from $\hat{\boldsymbol{Q}}$ in the following manner:
\begin{equation}
 \bar Q = diag\left ( \hat q_1, \hat q_2, \cdot \cdot \cdot , \hat q_{\left ( f-1 \right )}   \right )
\end{equation}
Modeled pulse signal, $\hat{\boldsymbol{X}_e}$ is obtained by:
\begin{equation}
 \hat{\boldsymbol{X}_e} = \mathbf{\hat{A}} \Phi
\end{equation}
An example of the modeled pulse signal is shown in Figure \ref{fig:Post_processing_steps}(d) along with its corresponding pulse spectrum in Figure \ref{fig:Post_processing_steps}(e). It can be observed from the Figures \ref{fig:Post_processing_steps}(c) and \ref{fig:Post_processing_steps}(e) that the HR estimation can be improved significantly after incorporating the proposed pulse modeling.

\subsection{HR tracking}
\begin{figure*}	
\centering
		{
		\includegraphics[scale = 0.35]{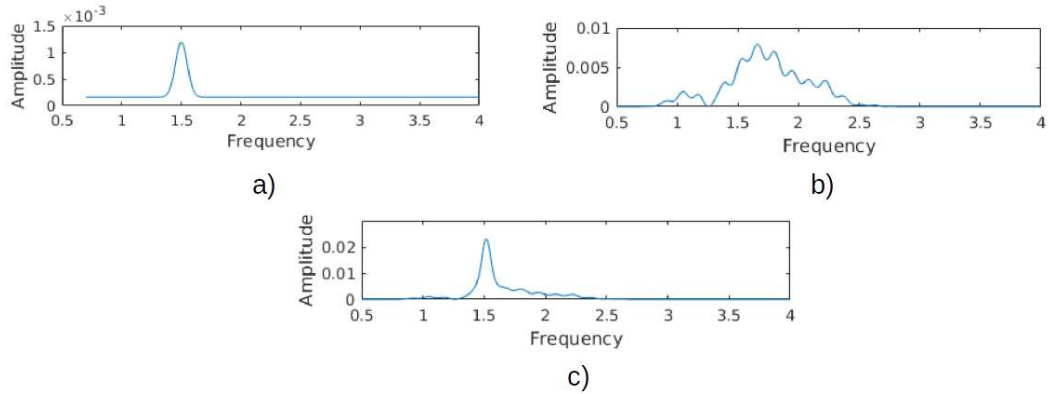}
	}
  \caption{Example of pulse spectrum during tracking: a) Prior, $P_e$ (for $h_i$ = 1.5 and $\frac{c}{q_i}$ = 0.05); b) Likelihood function, $P_l$; and c) Posterior, $P_P$. These depict the HR frequency (in beats per second) and their corresponding amplitude using X and Y-axes respectively.}
  \label{fig:Distribution_vis}
\end{figure*}

The spectrum of noise-free pulse signal should contain maximum amplitude at the HR frequency, but it is violated when the pulse signal contains noise. The modeled pulse signal obtained after applying the proposed pulse modeling technique, contains noise and thus, it can provide spurious HR. Usually, small HR change is observed between subsequent HR estimates. It motivates us to introduce the Bayesian framework which rectifies the spurious HR estimates. The framework consolidates the likelihood information derived from the current window and the prior information related to previously analyzed windows \cite{duda2012pattern}.

To leverage the observation that there is a small HR change between subsequent HR estimates, we want to define our prior information such that the large confidence value is provided when the fluctuation between the current HR and previous HR is small while a small confidence value is provided when the fluctuation is large. Furthermore, we need to provide low prior information about the previous HR whenever they are spurious, otherwise the error can be propagated in the subsequent HR estimates. An important characteristic of spurious HR is that its corresponding pulse spectrum contains multiple peaks, thus the spectrum has low PSNR \cite{yu2006method}. These conditions are met by defining the prior information for $ \left (i+1  \right )^{th}$ window using:
\begin{equation}
\label{prior_1}
P_g\left ( \boldsymbol{\theta} \right ) = \sqrt{\frac{\gamma_i}{{2\pi c}}} exp\left [ -\frac{1}{2}\left (   \frac{\gamma_i}{c}\left (\boldsymbol{\theta}-h_i  \right )   \right )^2 \right ]\sim \mathcal{N}\left ( h_i,\frac{c}{\gamma_i} \right )
\end{equation}
where $h_i$ and $\gamma_i$ denote the HR frequency and PSNR of the modeled pulse spectrum in the previous window (that is, $i^{th}$ window) respectively; $c$ is a predefined constant; $\boldsymbol{\theta}$ is the set of all probable HR frequencies; and $\mathcal{N}$ denote the normal distribution. It can be observed that the normal distribution is used in Equation \eqref{prior_1} such that mean and variance are given by $h_i$ and $\frac{c}{\gamma_i}$ respectively. Thus, small PSNR results in high variance which in turn results in low prior knowledge. Also, it is suggested by \cite{zhang2015troika} that the fluctuations between subsequent HR estimates usually lie within the range of -11bpm to +11bpm. Thus, we set $c$ equal to 4, so that 3 times of the variance covers most of our permissible HR estimates.

Our definition of prior information in Equation \eqref{prior_1} prohibits large fluctuation from the previous HR. Hence, if previous HR is spurious with low PSNR value, then the current HR values should be restricted with large range. But one should consider all the plausible HR frequencies when previous HR is spurious. To incorporate this intuition, we add a constant value in all the plausible HR frequencies, which are lying between 0.7 to 4Hz. That is, we modify the prior information using:
\begin{equation}
\label{prior_info_main}
P_e\left ( \boldsymbol{\theta} \right ) =  
    \begin{cases}
      \frac{P_u\left ( \boldsymbol{\theta} \right )+ P_g\left ( \boldsymbol{\theta} \right )}{ \left ( \sum_{\boldsymbol{\theta}=0.7}^{4}\left (P_u\left ( \boldsymbol{\theta} \right )+ P_g\left ( \boldsymbol{\theta} \right )  \right ) \right )     },
 & \text{if}\ 0.7<\theta<4Hz  \\
      0, & \text{otherwise}
    \end{cases}
    \end{equation}
where $P_e$ denotes the modified prior information; $P_g$ is the distribution described in Equation \eqref{prior_1}; and $P_u$ is given by:
\begin{equation}
\label{chat}
P_u\left ( \boldsymbol{\theta} \right ) =  
    \begin{cases}
      \hat{c}, & \text{if}\ 0.7<\boldsymbol{\theta}<4Hz  \\
      0, & \text{otherwise}
    \end{cases}
\end{equation}
In essence, $P_u$ is the uniform distribution, defined in the HR frequency ranges of 0.7 to 4 Hz such that any frequency is equally probable with the value of $\hat{c}$. We describe the parameter selection of $\hat{c}$ in Section \ref{Experimental_Results}. Furthermore, when the first window is analyzed $P_g$ is set to zero for all the possible HR frequency ranges, so that all the values are equally likely and hence no useful prior information is utilized.

\begin{figure}	
  \centering
  \includegraphics[width=0.9\linewidth]{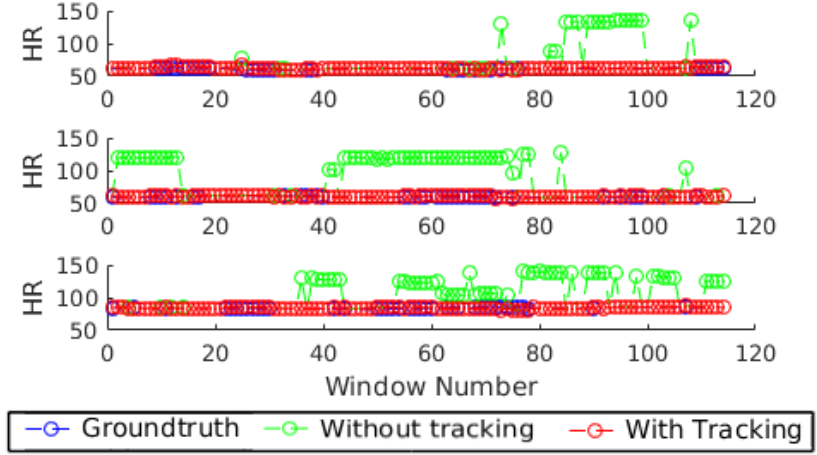}
  \caption{Examples of HR monitoring with and without tracking. X-axis denotes the window number and y-axis denotes the corresponding HR (in bpm). (Figures best viewed in colors)}
  \label{fig:tracking_g}
\end{figure}

The likelihood function is denoted by $P_l\left ( \boldsymbol{S_{i+1}}|\boldsymbol{\theta} \right )$ where $\boldsymbol{S_{i+1}}$ denotes the spectrum of reconstructed pulse signal $\hat{\boldsymbol{X}_e}$ corresponding to the $ \left (i+1  \right )^{th}$ window. We estimate it using:
\begin{equation}
\label{posteriori}
 P_l\left ( \boldsymbol{S_{i+1}}|\boldsymbol{\theta} \right ) = \frac{\boldsymbol{S_{i+1}}\left ( \boldsymbol{\theta} \right )}{\left (\sum_{\boldsymbol{\theta}=0.7}^{4Hz}\boldsymbol{S_{i+1}}\left ( \boldsymbol{\theta} \right )  \right )}
 \end{equation}
The posterior probability, $P_p\left ( \boldsymbol{\theta}|\boldsymbol{S_{i+1}} \right )$ is evaluated by applying the Bayes rule \cite{duda2012pattern}, that is,
\begin{equation}
\label{bayes}
P_p\left ( \boldsymbol{\theta}|\boldsymbol{S_{i+1}} \right ) = \frac{ P_l\left ( \boldsymbol{S_{i+1}}|\boldsymbol{\theta} \right ) P_e\left ( \boldsymbol{\theta} \right )}{P\left ( \boldsymbol{S_{i+1}} \right )}
\end{equation}
where $P\left ( \boldsymbol{S_{i+1}} \right )$ is the evidence factor. Equations \eqref{prior_info_main}, \eqref{posteriori} and \eqref{bayes} can be combined in the following manner:
\begin{equation}
P_p\left ( \boldsymbol{\theta}|\boldsymbol{S_{i+1} }\right ) = \frac{\boldsymbol{S_{i+1}}\left ( \boldsymbol{\theta} \right )\left ( P_u\left ( \boldsymbol{\theta} \right )+ P_g\left ( \boldsymbol{\theta} \right )\right ) }{Z_2}
\end{equation}
where $Z_2$ is a normalization coefficient given by:
\begin{equation}
Z_2 =  \left ( \sum_{\boldsymbol{\theta}=0.7}^{4}\left (P_u\left ( \boldsymbol{\theta} \right )+ P_g\left ( \boldsymbol{\theta} \right )
\right ) \right ) P\left ( \boldsymbol{S_{i+1}} \right ) \left (\sum_{\boldsymbol{\theta}=0.7}^{4Hz}\boldsymbol{S_{i+1}}\left ( \boldsymbol{\theta} \right )  \right ) 
\end{equation}
An illustration of the prior information, likelihood function and their corresponding posterior probability is shown in Figure \ref{fig:Distribution_vis}. We apply maximum a posteriori estimation for HR frequency estimation which provides the minimum-error-rate classifier based on zero-one loss function \cite{duda2012pattern}. For brevity, the expected loss incurred on selecting a particular frequency, $d$ is given by:
\begin{equation}
\label{eq1}
 R\left ( \boldsymbol{\theta}=d|\boldsymbol{S_{i+1}} \right ) = \sum_{K = 0.7}^{4}L\left ( d,\boldsymbol{\theta} \right )P_p\left ( \boldsymbol{\theta} = K|\boldsymbol{S_{i+1}} \right )
\end{equation}
where $R$ is the incurred loss and $L$ represent the loss function given by:
\begin{equation}
\label{eq2}
 L\left ( d,\boldsymbol{\theta} \right ) = 
    \begin{cases}
      0, & \text{if}\ \boldsymbol{\theta} = d  \\
      1, & \text{otherwise}
    \end{cases}
\end{equation}
Further, it is obvious that the sum of likelihood function at all the possible values (which lies between 0.7 to 4Hz in our case) will be equal to one, that is,
\begin{equation}
\label{eq3}
 \sum_{K=0.7}^{4}P_p\left ( \boldsymbol{\theta}=K|\boldsymbol{S_{i+1}} \right ) = 1
\end{equation}
It can be seen by combining Equations \eqref{eq1}, \eqref{eq2}  and \eqref{eq3} that:
\begin{equation}
 R\left ( \boldsymbol{\theta}=d|\boldsymbol{S_{i+1}} \right ) = 1-P_p\left ( \boldsymbol{\theta}=d|\boldsymbol{S_{i+1}} \right )
\end{equation}
Hence, expected loss, $R$ is minimized when $\boldsymbol{\theta}$ is set to the value that maximizes the posterior probability $P_p$, that is, $\boldsymbol{\theta}$ is set to HR frequency. Hence, we obtain the HR frequency corresponding to $ \left (i+1  \right )^{th}$ window, $ h_{i+1}$ using:
\begin{equation}
 {h_{i+1}}={\underset {\boldsymbol{\theta} }{\operatorname {arg\,max} }}\ P_p\left ( \boldsymbol{\theta}|\boldsymbol{S_{i+1}} \right )
\end{equation}
The corresponding HR is given by:
\begin{equation}
 \hat {H} \left (i+1  \right ) = round\left ( { {h}_{i+1}} \times 60\right )
\end{equation}
where $round$ operator rounds off the value to the nearest integer. Some examples depicting the usefulness of the proposed HR tracking are shown in Figure \ref{fig:tracking_g}. The figure depicts the actual HR monitoring along with the predicted HR monitoring when the proposed HR tracking is avoided and utilized. It demonstrates that the HR monitoring can be improved significantly when the proposed HR tracking is used.

\section{Experimental Results}     \label{Experimental_Results}
\subsection{Data Recording of Our Dataset}
\begin{figure}	
		{
		\includegraphics[scale = 0.35]{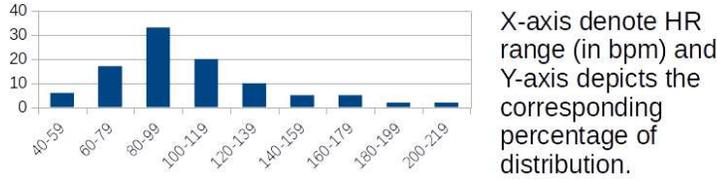}
	}
  \caption{Distribution of HR in the database. }
  \label{fig:Distribution}
\end{figure}

The performance of our method, $MOMBAT$ is evaluated on Intel i5-2400 CPU 3.10 GHz. Total 65 face videos have been collected from 65 different subjects (34 males and 31 females), out of which 15 videos are used for parameter selection (or training) and the remaining 50 videos are used for performance evaluation (or testing). The videos are acquired from Logitech webcam C270 camera which is mounted on a laptop and the subjects are free to perform natural facial movements and head pose variations. The resolution of these acquired videos is 640$\times$480 pixels. Furthermore, we avoid any compression mechanism and save the videos in AVI raw format. These are acquired for 1 minute at 30 frames per second. The ground truth is obtained by simultaneously acquiring the actual pulse from the right index fingertip using CMS 50D+ pulse oximeter. The percentage of distribution of ground truth HR estimation from the acquired database is shown in Figure \ref{fig:Distribution}.

\subsection{Performance Measurement} \label{PerformanceMeasurement}

The performance metrics used in our experiments are based on the predicted predicted HR error, $\left ( \bar P\left ( i,j \right )-\bar A\left ( i,j \right ) \right )$, where $\bar P\left ( i,j \right )$ and $\bar A\left ( i,j \right )$ denote the predicted and actual HR estimates respectively for $i^{th}$ subject in $j^{th}$ window. Accurate HR monitoring method requires that the prediction error is close to zero, alternatively, the mean $\mu $ and standard deviation $\sigma$ of the prediction error should be close to zero. Likewise, the percentage of samples with absolute error less than 5 bpm, $err_5$ should be close to 100\% for correct HR monitoring. Another metric employed for the evaluation is mean average error, $MAE$ of all the subjects which is given by:
\begin{equation}
 MAE = \frac{\sum_{i=1}^{z}\sum_{j=1}^{n_i}\left | \bar P\left ( i,j \right ) - \bar A\left ( i,j \right ) \right |}{\sum_{i=1}^{z}n_i}
\end{equation}
where $\left | \bullet \right |$ is the absolute operator; $n_i$ represents the number of windows for $i_{th}$ subject; and $z$ is the total number of subjects. Lower value of $MAE$ indicates that the predicted and estimated HR estimates are close to each other. Similarly, we also use total time, $t_s$ required for HR monitoring in seconds as a performance metric. Furthermore, we used the Pearson correlation coefficient, $\rho$ to evaluate the similarity between two variables in terms of linear relationship. It lies between -1 to 1 and is given by:
\begin{equation}
 \rho={\frac {\operatorname {cov} (\bar P,\bar A)}{\sigma \left ( \bar P \right ) \times \sigma \left ( \bar A \right )}}
\end{equation}
where $cov$ and $\sigma$ are the covariance and standard deviation operator respectively. Better HR estimation requires high similarity between the predicted and actual HR, that is, high $\rho$.

\begin{figure}	
		{
		\includegraphics[scale = 0.35]{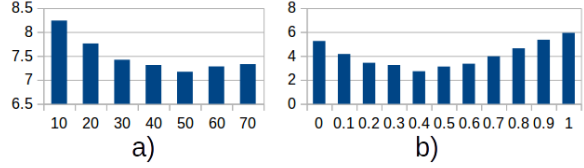}
	}
  \caption{Performance analysis for: a) $\alpha$ and b) $\hat{c}$. X and Y-axes denote the parameter value and $MAE$ (in bpm) respectively.}
  \label{fig:parameter}
\end{figure}

\subsection{Parameter Selection}    \label{ParameterSelection}

Our proposed method $MOMBAT$ requires proper selection of four parameters, which are: i) window size; ii) overlapping window size; iii) $\alpha$  representing the number of basis in pulse modeling; and iv) $\hat{c}$ which is required for defining prior in HR tracking (refer Equation \eqref{chat}). Since minimum possible heart beats can be 42bpm and the proper window should contain at least two cycles of the cardiovascular pulse, we selected the window of 4 second video where slightly more than two cycles can be observed. Similarly, overlap between the successive windows is chosen in an application specific manner. We focus on frequently updating the previous HR from the new HR, typically, twice in a second. Hence, we set the overlap between successive windows at 3.5 seconds. The remaining parameters are set by the value, providing minimum $MAE$ on the training set. The $MAE$ for different parameter values are shown in Figure \ref{fig:parameter}. We set $\alpha$ and $\hat{c}$ equal to 50 and 0.4 respectively, where minimum $MAE$ is attained on the training set.

\begin{table}
\begin{threeparttable}
\caption{Description of subversions of $MOMBAT$}
{      \begin{tabular}{ |p{2.2cm}||p{1.8cm}|p{1.8cm}|p{1.8cm}|p{1.8cm}| }
 \hline
Method    & Color  & Image        & Pulse             & Bayesian \\
          &channel & registration & modeling\tnote{1} & tracking \\
 \hline
$NorSysR$     & Red   & No  & NA & No  \\
$NorSysI$     & Green & No  & NA & No  \\
$NorSys$      & Green & Yes & NA         & No\\
$PulseModP$   & Green & Yes & Polynomial & No\\
$PulseModL$   & Green & Yes & Legendre   & No\\
$PulseMod$    & Green & Yes & Fourier    & No\\
$BayTrack$    & Green & Yes & NA         & Yes\\
\textbf{$\boldsymbol{MOMBAT}$}      &  \textbf{Green} & \textbf{Yes} & \textbf{Fourier} & \textbf{Yes}\\
 \hline
\end{tabular}
}
\begin{tablenotes}
\item[1] NA means pulse modeling is avoided. Otherwise, the basis name is mentioned.
\end{tablenotes}
\label{tab:subversions}
\end{threeparttable}
\end{table} 

\subsection{Performance Evaluation}   \label{PerformanceEvaluation}
For rigorous performance analysis, we create several other  methods from our proposed method, $MOMBAT$ by avoiding or replacing its components. The following methods are considered for the performance analysis: a) $NorSysI$ which is obtained by avoiding image registration, pulse modeling and HR tracking in $MOMBAT$; b) $NorSysR$ which is same as $NorSysI$ except that it uses red light instead of green light for extracting the temporal signals; c) $NorSys$ which is obtained by avoiding pulse modeling and HR tracking in $MOMBAT$; d) $PulseModP$ and $PulseModL$ which avoid HR tracking, but utilize polynomial and Legendre basis \cite{gupta2015fingerprint} respectively, instead of Fourier basis for pulse modeling in $MOMBAT$; e) $PulseMod$ which is given by considering proposed (or Fourier basis based) pulse modeling, but avoiding HR tracking in $MOMBAT$; and f) $BayTrack$ which is given by avoiding pulse modeling, but considering HR tracking from $MOMBAT$. The description of these subversions of $MOMBAT$ is provided in Table \ref{tab:subversions}. Furthermore, we compare our method with the following existing well known methods: \cite{balakrishnan2013detecting}; CHROM \cite{de2013robust}; POS \cite{wang2016algorithmic}; \cite{huang2016accurate}; \cite{tulyakov2016self}; \cite{rodriguez2018video}; \cite{qiu2018evm}; and \cite{gupta2018robust}. Pulse signal is extracted from the Lagrangian temporal signals using Principal Component Analysis in \cite{balakrishnan2013detecting}. \cite{huang2016accurate} registers the face and utilizes Eulerian temporal signals. Model based methods are utilized in \cite{de2013robust} and \cite{wang2016algorithmic} where temporal signals are extracted by fusing RGB color channels. Optical and physiological properties of skin reflection are used to perform such a fusion. Methods  \cite{balakrishnan2013detecting}, \cite{huang2016accurate}, \cite{de2013robust} and \cite{wang2016algorithmic} provide one HR value. To conform these methods with $MOMBAT$, we extract the window and then analyze each window using these methods for HR monitoring. We are unable to conduct the comparative analysis with \cite{li2014remote} for HR monitoring because it requires large window size as described in \cite{tulyakov2016self}. In \cite{qiu2018evm}, CNN trained on several windows is used for HR monitoring. The training and test sets contain different windows of the same subjects. For more rigorous analysis, we also perform the experimentation with another method $ModCNN$ where \cite{qiu2018evm} is used, except that its training and testing sets do not contain windows of the same subjects.

\subsection{Performance Analysis on Our Dataset}
Our experimental results on our dataset are presented in Table \ref{tab:Expermental_results}. It can be inferred from the table that \cite{balakrishnan2013detecting} provides the most spurious HR monitoring because it requires the tracking of facial features, which is easily affected by expressions. Likewise, \cite{huang2016accurate} exhibits lower performance than the other methods except  \cite{balakrishnan2013detecting} because it averages all the temporal signals for pulse extraction. This is error-prone because large noise in few temporal signals due to facial expressions, can tremendously affect the cardiovascular pulse after averaging. Both \cite{balakrishnan2013detecting} and \cite{huang2016accurate} employ highly time consuming feature tracking and BSS. In contrast, \cite{rodriguez2018video} extracts the pulse signal using only green channel intensity differences of full face and avoiding computationally expensive BSS step and feature tracking. It enables \cite{rodriguez2018video} to perform in the most computationally efficient manner, but such a method performs spuriously because it is easily affected by facial deformation, as shown in \cite{gupta2018robust}.

\begin{table}
\begin{threeparttable}
\caption{Comparative Results of HR Monitoring on our Dataset}
{      \begin{tabular}{ |p{2cm}||p{1.3cm}|p{1.2cm}|p{0.7cm}|p{1.3cm}|p{1.1cm}|p{0.8cm}|  }
 \hline
Method                              & $\mu $ & $\sigma$ & $err_5$ & $MAE$ & $\rho$ & $t_s$ \\
 \hline
\cite{balakrishnan2013detecting}    &  -18.4120 & 27.6195 & 38 & 22.5972 & -0.1793 & 25.72 \\
\cite{huang2016accurate}            &  -9.0745  & 20.2534 & 75 & 10.0305 &  0.2515 & 30.37\\
\cite{rodriguez2018video}           &  9.5317   & 21.0467 & 70 & 11.3885 & 0.3310  & 1.24\\
 CHROM \cite{de2013robust}      &    -8.1932 &  19.1917 &  76 &  9.843 &   0.3015 &   6.81\\
 POS \cite{wang2016algorithmic} &    -8.9827 &  19.7920 &  77 &  10.214 &   0.2912 &   6.81\\
 $NorSysR$       &    -10.8246 &  21.4921 &  72 &  10.946 &   0.2847 &   6.80\\
$NorSysI$                           &   -6.9634 & 18.7418 & 80 &  8.7382 &  0.3106 &  6.80\\
\cite{tulyakov2016self}             &  6.8242   & 18.3521 & 81 & 8.1864  & 0.4256  & 19.41\\
$NorSys$                            &  -6.4405  & 17.4389 & 83 & 7.3813  & 0.4486  & 6.84\\
 $PulseModP$                        &  -6.4079  & 17.3964 & 82 & 7.2503  &  0.4504 &  9.63\\
 $PulseModL$                        &  -6.3865  &  17.3726& 83 & 7.2057  & 0.4542  &  9.63\\
$PulseMod$                          &   -6.1783 & 17.1010 & 85 & 7.1853  & 0.4627  &  9.63\\
$BayTrack$                          &  -0.5864  & 5.7052  & 94 & 2.2154  & 0.8821  &  7.06\\
\cite{qiu2018evm}                   &  0.9275   & 7.3472  & 90 & 3.4588  & 0.8226  &  9.92\\
$ModCNN$                            & -10.1539  & 21.6781 & 31 & 18.2169 & -0.0337 &  9.92\\
\cite{gupta2018robust}              &  0.4667 & 4.8230    & 89 & 2.4968  & 0.8601  &  16.81\\
\textbf{$\boldsymbol{MOMBAT}$}      &  \textbf{-0.1041} & \textbf{2.6172} & \textbf{97} & \textbf{1.3293} & \textbf{0.9746} & \textbf{9.78}\\
 \hline
\end{tabular}
}
\begin{tablenotes}
\item[]  Unit of: i) $err_5$ is $\%$; ii) 
$\mu $, $\sigma$ and $MAE$ is bpm; and $t_s$ is seconds.
\end{tablenotes}
\label{tab:Expermental_results}
\end{threeparttable}
\end{table} 

$NorSysR$ and $NorSysI$ are different only in the way that they utilize red and green light respectively for the temporal signal extraction. It can be observed from Table \ref{tab:Expermental_results} that $NorSysI$ performs better than $NorSysR$, which indicates that green light is more effective in photo-plethysmographic imaging than red light. This observation is also mentioned in \cite{verkruysse2008remote}. Furthermore, $NorSysI$ performs better than CHROM \cite{de2013robust} and POS \cite{wang2016algorithmic}, which utilize optical and physiological properties of skin reflection to consolidate RGB color channels for temporal signal extraction. It is because CHROM and POS do not provide correct HR estimation when pulse signal and noise share similar amplitudes \cite{wang2016algorithmic}. In addition, POS fails when face videos are illuminated by multiple light sources \cite{wang2016algorithmic}.

$NorSysI$ is the same as $NorSys$ except that $NorSysI$ avoids image registration and Table \ref{tab:Expermental_results} points out that $NorSys$ performs better than $NorSysI$. It indicates that performance can be increased by utilizing image registration. Average computational time of $NorSysI$ and $NorSys$ are 6.8 second and 6.84 second, respectively, out of which, BSS is the most computationally expensive step requiring 5.13 seconds. Method $NorSys$ also performs better than \cite{tulyakov2016self} due to better ROI selection, image registration and proper BSS technique. \cite{tulyakov2016self} utilizes matrix completion to mitigate the noise, which increases the computation time significantly. Also, it can be observed from the table that $PulseModP$, $PulseModL$, $PulseMod$ and $BayTrack$ perform better HR monitoring than $NorSys$. $PulseModP$, $PulseModL$ and $PulseMod$ mitigate the problems of out-of-plane movements in $NorSys$ by modeling the pulse signal (refer Figure \ref{fig:Post_processing_steps} for example) and in return, they incur an additional average time of 0.31 and 2.48 sec for the modeling and quality estimation respectively. $PulseMod$ performs better than $PulseModL$ and $PulseModP$ which demonstrate that Fourier basis is better suited for pulse signal modeling. Similarly, $BayTrack$ performs better monitoring than $NorSys$ because it rectifies the HR estimates by incorporating the prior knowledge of the HR estimates. Some of its examples are shown in Figure \ref{fig:tracking_g}. It incurs an additional average time of 0.22 sec than $NorSys$ due to PSNR estimation.
 
\begin{figure}	
	{
		\includegraphics[scale = 0.5]{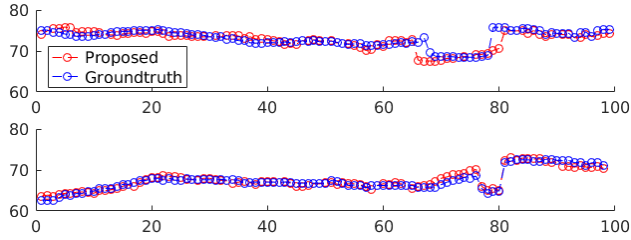}
	}	 
  \caption{HR Monitoring by Our $MOMBAT$ under Large HR Fluctuations. X and Y-axes denote the window number and HR (in bpm) respectively. (Figures best viewed in colors)}
  \label{fig:SuccessResult}
\end{figure}

\begin{figure}	
	{
		\includegraphics[scale = 0.5]{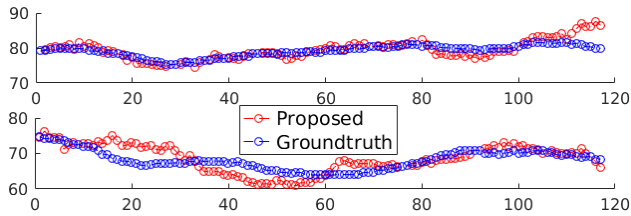}
	}	
  \caption{Erroneous HR Monitoring by Our $MOMBAT$. X and Y-axes denote the window number and HR (in bpm) respectively. (Figures best viewed in colors)}
  \label{fig:errorResult}
\end{figure}

 Table \ref{tab:Expermental_results} indicates that \cite{qiu2018evm} exhibits good HR monitoring when the training and test sets contain different windows of the same subjects. But when the training and testing sets do not contain windows of the same subjects (that is, $ModCNN$) then there is significant performance degradation. It points out that CNN employed by \cite{qiu2018evm}, leverages the facial texture and skin color for HR monitoring. This is obviously a wrong way of performing HR monitoring. Likewise, \cite{gupta2018robust} relying on only face reconstruction is incompetent to handle out-of-plane movements and hence, provide spurious HR estimates. But our method, $MOMBAT$ handles most of the spurious cases by utilizing the pulse modeling and Bayesian tracking. Furthermore, it provides the best HR monitoring  amongst all the methods.But it incurs an additional average time of 2.94 sec when compared with $NorSys$ due to modeling and tracking. Such small time differences can be neglected to achieve significantly better HR monitoring for the face videos of 54 sec. A few cases where $MOMBAT$ has successfully performed the HR monitoring are shown in Figure \ref{fig:SuccessResult}. Just like other existing methods, $MOMBAT$ may perform spuriously when the face video contains noise that persists for long duration. Some such spurious monitoring cases by $MOMBAT$ are shown in Figure \ref{fig:errorResult}.

\subsection{Performance Analysis on COHFACE}   \label{CohfaceAck}
 
One major factor that hampers the progress in this realm of HR analysis using face video is the lack of appropriate datasets \cite{chen2018video}. It is stated by \cite{chen2018video} that several existing publicly available datasets that are extensively used in the literature, are not appropriate for HR estimation using face videos. One such example is MAHNOB-HCI dataset \cite{soleymani2011multimodal} which involves negligible illumination variations induced by the movie and thus, unable to cater complex real-world scenarios. COHFACE dataset is regarded as a more challenging dataset to cater more realistic conditions than MAHNOB-HCI by \cite{heusch2017reproducible}. Thus, we have conducted our experiments on COHFACE dataset as well. But it lacks significant motion variations and thus, we create and conduct experiments on our dataset for better evaluation of our proposed method, $MOMBAT$.

\begin{table}
\begin{threeparttable}
\caption{Comparative Results of HR Monitoring on COHFACE}
{      \begin{tabular}{ |p{2cm}||p{1.3cm}|p{1.2cm}|p{0.7cm}|p{1.3cm}|p{1.1cm}|p{0.8cm}|  }
 \hline
Method                              & $\mu $ & $\sigma$ & $err_5$ & $MAE$ & $\rho$  \\
 \hline
\cite{balakrishnan2013detecting}    &  -10.1664 & 19.2326 & 70 & 12.3612 & 0.0897  \\
\cite{huang2016accurate}            &   9.3481  & 18.5818 & 74 & 11.2493 & 0.1264 \\
\cite{rodriguez2018video}      &  6.8215 & 19.4267& 67 & 16.3568& -0.1129 \\
 CHROM \cite{de2013robust}      &    -7.9135 &  19.1191 &  75 &  10.384 &   0.1315\\
 POS \cite{wang2016algorithmic} &    -7.1263 &  18.3902 &  77 &  9.621 &   0.1617\\
 $NorSysR$          &    -7.8460 &  18.9212 &  72 &  10.696 &   0.1248\\
$NorSysI$                          &  -6.2179  & 16.3592 & 80 & 9.0424 & 0.1721 \\
\cite{tulyakov2016self}             &  6.9628   & 15.6330 & 81 & 8.9354& 0.1813 \\
$NorSys$                            &  -5.3672  & 15.8627 & 83 & 8.7956 & 0.2065 \\
 $PulseModP$                        &  -4.6730  & 14.7966 & 85 & 8.2718 & 0.2108 \\
 $PulseModL$                        &  -4.5203  & 14.7171 & 85 & 8.1629 & 0.2256  \\
$PulseMod$                          &  -4.0942  & 13.9043 & 86 & 7.9250 & 0.2614 \\
$BayTrack$                     & -1.0576 & 9.4852 & 90 & 6.4797 & 0.5352 \\
\cite{qiu2018evm}              &  2.3874 & 11.9163& 89 & 6.8128 & 0.5036 \\
$ModCNN$          & -4.3561 & 18.2781& 56 & 14.8243& -0.0696 \\
\cite{gupta2018robust}         &  1.4666 & 12.6595& 88 & 6.5411 & 0.5252 \\
\textbf{$\boldsymbol{MOMBAT}$}       &  \textbf{-0.9832} & \textbf{7.3823} & \textbf{92} & \textbf{5.8923} & \textbf{0.6184} \\
 \hline
\end{tabular}
}
\begin{tablenotes}
\item[] Unit of $err_5$ is $\%$ while unit of 
$\mu $, $\sigma$ and $MAE$ is bpm.
\end{tablenotes}
\label{tab:Expermental_results_2}
\end{threeparttable}
\end{table} 

The COHFACE dataset contains 160 face videos acquired from 40 subjects. Experiments are conducted on this dataset using the performance metrics, parameter selection and methods described in Section \ref{PerformanceMeasurement}, Section \ref{ParameterSelection} and Section \ref{PerformanceEvaluation} respectively. The corresponding results are shown in Table \ref{tab:Expermental_results_2}. It can be observed that these results are similar to the results on our dataset, that is, $MOMBAT$ performs best amongst the considered methods. Furthermore, it can be observed that the efficacy of $MOMBAT$ reduces slightly when COHFACE dataset is considered rather than our dataset. It is because the COHFACE dataset contains compressed videos which deteriorate the HR analysis \cite{rapczynski2019effects}.

\section{Conclusions}                     \label{conclusion}

This paper has proposed an HR monitoring method, $MOMBAT$, that is, MOnitoring using Modeling and BAyesian Tracking. It has utilized the face videos acquired from a low cost camera in contact-less manner, for HR monitoring. HR monitoring using face videos can be error-prone due to facial expressions, out-of-plane movements, camera parameters and environmental factors. Thus, our $MOMBAT$ have alleviated these issues to improve the HR monitoring by introducing pulse modeling and Bayesian HR tracking. The proposed Fourier basis based modeling mitigates the out-of-plane movements by successfully reconstructing the poor quality pulse signal estimates using the good quality pulse signal estimates. The noise can result in some spurious HR estimates, but our proposed Bayesian decision theory based HR tracking mitigates such cases to improve the HR monitoring.

Experimental results have demonstrated that HR monitoring can be significantly improved when both pulse modeling and HR tracking are incorporated. Further, they have indicated that our $MOMBAT$ perform the monitoring in near real-time, with an average absolute error of 1.3293 bpm and the Pearson correlation of 0.9746 between predicted and actual HR. This indicates that our method, $MOMBAT$ can be effectively used for HR monitoring.

Our method $MOMBAT$ can perform spuriously when the face video contains motion that persists for long duration. Our future work will investigate the possibilities to handle this issue by fusing it with Lagrangian techniques. Our method requires some parameter selection. Amongst them, the number of basis depends on the sampling rate. We will be collecting a larger database at different sampling rates using different video compression techniques. It will be used to explore the efficacy of convolutional neural networks \cite{qiu2018evm} and mitigate the compression artifacts for better HR analysis.

\section*{Acknowledgement}
Our method was tested on a publicly available dataset COHFACE Dataset in Section \ref{CohfaceAck}. It was provided by the Idiap Research Institute, Martigny, Switzerland.

\bibliography{clean}

\begin{thebibliography}{10}

\bibitem{balakrishnan2013detecting}
Guha Balakrishnan, Fredo Durand, and John Guttag.
\newblock Detecting pulse from head motions in video.
\newblock In {\em IEEE Conference on Computer Vision and Pattern Recognition
  (CVPR)}, pages 3430--3437, 2013.

\bibitem{baltruvsaitis20123d}
Tadas Baltru{\v{s}}aitis, Peter Robinson, and Louis-Philippe Morency.
\newblock 3d constrained local model for rigid and non-rigid facial tracking.
\newblock In {\em IEEE Conference on Computer Vision and Pattern Recognition
  (CVPR)}, pages 2610--2617. IEEE, 2012.

\bibitem{baltrusaitis2013constrained}
Tadas Baltrusaitis, Peter Robinson, and Louis-Philippe Morency.
\newblock Constrained local neural fields for robust facial landmark detection
  in the wild.
\newblock In {\em IEEE International Conference on Computer Vision Workshops
  (ICCV-W)}, pages 354--361, 2013.

\bibitem{baltruvsaitis2016openface}
Tadas Baltru{\v{s}}aitis, Peter Robinson, and Louis-Philippe Morency.
\newblock Openface: an open source facial behavior analysis toolkit.
\newblock In {\em IEEE Winter Conference on Applications of Computer Vision
  (WACV)}, pages 1--10. IEEE, 2016.

\bibitem{berntson1997heart}
Gary~G Berntson, J~Thomas~Bigger, Dwain~L Eckberg, Paul Grossman, Peter~G
  Kaufmann, Marek Malik, Haikady~N Nagaraja, Stephen~W Porges, J~Philip Saul,
  Peter~H Stone, et~al.
\newblock Heart rate variability: origins, methods, and interpretive caveats.
\newblock {\em Psychophysiology}, 34(6):623--648, 1997.

\bibitem{Broj2}
Brojeshwar Bhowmick, Apurbaa Mallik, and Arindam Saha.
\newblock Mobiscan3d: A low cost framework for real time dense 3d
  reconstruction on mobile devices.
\newblock In {\em Intl Conf on Ubiquitous Intelligence and Computing}, pages
  783--788, 2014.

\bibitem{borghi2018face}
Guido Borghi, Matteo Fabbri, Roberto Vezzani, Rita Cucchiara, et~al.
\newblock Face-from-depth for head pose estimation on depth images.
\newblock {\em IEEE transactions on pattern analysis and machine intelligence},
  2018.

\bibitem{camm1996heart}
A~John Camm, Marek Malik, J~Thomas Bigger, G{\"u}nter Breithardt, Sergio
  Cerutti, Richard~J Cohen, Philippe Coumel, Ernest~L Fallen, Harold~L Kennedy,
  RE~Kleiger, et~al.
\newblock Heart rate variability: standards of measurement, physiological
  interpretation and clinical use. task force of the european society of
  cardiology and the north american society of pacing and electrophysiology.
\newblock 1996.

\bibitem{Broj3}
Kingshuk Chakravarty, Suraj Suman, Brojeshwar Bhowmick, Aniruddha Sinha, and
  Abhijit Das.
\newblock Quantification of balance in single limb stance using kinect.
\newblock In {\em IEEE International Conference on Acoustics, Speech and Signal
  Processing}, pages 854--858, 2016.

\bibitem{chen2018video}
Xun Chen, Juan Cheng, Rencheng Song, Yu~Liu, Rabab Ward, and Z~Jane Wang.
\newblock Video-based heart rate measurement: Recent advances and future
  prospects.
\newblock {\em IEEE Transactions on Instrumentation and Measurement},
  68(10):3600--3615, 2019.

\bibitem{de2013robust}
Gerard De~Haan and Vincent Jeanne.
\newblock Robust pulse rate from chrominance-based rppg.
\newblock {\em IEEE Transactions on Biomedical Engineering}, 60(10):2878--2886,
  2013.

\bibitem{decarlo1997meaning}
Lawrence~T DeCarlo.
\newblock On the meaning and use of kurtosis.
\newblock {\em Psychological methods}, 2(3):292, 1997.

\bibitem{duda2012pattern}
Richard~O Duda, Peter~E Hart, and David~G Stork.
\newblock {\em Pattern classification}.
\newblock John Wiley \& Sons, 2012.

\bibitem{gupta2017accurate}
Puneet Gupta, Brojeshwar Bhowmick, and Arpan Pal.
\newblock Accurate heart-rate estimation from face videos using quality-based
  fusion.
\newblock In {\em IEEE International Conference on Image Processing, (ICIP)},
  pages 4132--4136. IEEE, 2017.

\bibitem{gupta2017serial}
Puneet Gupta, Brojeshwar Bhowmick, and Arpan Pal.
\newblock Serial fusion of eulerian and lagrangian approaches for accurate
  heart-rate estimation using face videos.
\newblock In {\em IEEE International Conference of the Engineering in Medicine
  and Biology Society (EMBC)}, pages 2834--2837. IEEE, 2017.

\bibitem{gupta2018exploring}
Puneet Gupta, Brojeshwar Bhowmick, and Arpan Pal.
\newblock Exploring the feasibility of face video based instantaneous
  heart-rate for micro-expression spotting.
\newblock In {\em IEEE Conference on Computer Vision and Pattern Recognition
  Workshops (CVPRW)}, pages 1316--1323, 2018.

\bibitem{gupta2018robust}
Puneet Gupta, Brojeshwar Bhowmik, and Arpan Pal.
\newblock Robust adaptive heart-rate monitoring using face videos.
\newblock In {\em IEEE Winter Conference on Applications of Computer Vision
  (WACV)}, pages 530--538. IEEE, 2018.

\bibitem{gupta2015accurate}
Puneet Gupta and Phalguni Gupta.
\newblock An accurate finger vein based verification system.
\newblock {\em Digital Signal Processing}, 38:43--52, 2015.

\bibitem{gupta2015fingerprint}
Puneet Gupta and Phalguni Gupta.
\newblock Fingerprint orientation modeling using symmetric filters.
\newblock In {\em IEEE Winter Conference on Applications of Computer Vision
  (WACV)}, pages 663--669. IEEE, 2015.

\bibitem{gupta2016accurate}
Puneet Gupta and Phalguni Gupta.
\newblock An accurate fingerprint orientation modeling algorithm.
\newblock {\em Applied Mathematical Modelling}, 40(15):7182--7194, 2016.

\bibitem{heusch2017reproducible}
Guillaume Heusch, Andr{\'e} Anjos, and S{\'e}bastien Marcel.
\newblock A reproducible study on remote heart rate measurement.
\newblock {\em arXiv preprint arXiv:1709.00962}, 2017.

\bibitem{huang2016accurate}
Chong Huang, Xin Yang, and Kwang-Ting~Tim Cheng.
\newblock Accurate and efficient pulse measurement from facial videos on
  smartphones.
\newblock In {\em IEEE Winter Conference on Applications of Computer Vision
  (WACV)}, pages 1--8. IEEE, 2016.

\bibitem{huang2016self}
Ming-Chun Huang, Jason~J Liu, Wenyao Xu, Changzhan Gu, Changzhi Li, and Majid
  Sarrafzadeh.
\newblock A self-calibrating radar sensor system for measuring vital signs.
\newblock {\em IEEE transactions on biomedical circuits and systems},
  10(2):352--363, 2016.

\bibitem{huber1985projection}
Peter~J Huber.
\newblock Projection pursuit.
\newblock {\em The annals of Statistics}, pages 435--475, 1985.

\bibitem{lam2015robust}
Antony Lam and Yoshinori Kuno.
\newblock Robust heart rate measurement from video using select random patches.
\newblock In {\em International Conference on Computer Vision (ICCV)}, pages
  3640--3648, 2015.

\bibitem{lee2015heart}
Dongseok Lee, Jeehoon Kim, Sungjun Kwon, and Kwangsuk Park.
\newblock Heart rate estimation from facial photoplethysmography during dynamic
  illuminance changes.
\newblock In {\em IEEE International Conference of the Engineering in Medicine
  and Biology Society (EMBC)}, pages 2758--2761. IEEE, 2015.

\bibitem{li2014remote}
Xiaobai Li, Jie Chen, Guoying Zhao, and Matti Pietikainen.
\newblock Remote heart rate measurement from face videos under realistic
  situations.
\newblock In {\em IEEE Conference on Computer Vision and Pattern Recognition
  (CVPR)}, pages 4264--4271, 2014.

\bibitem{Broj4}
Apurbaa Mallik, Brojeshwar Bhowmick, and Shahnawaz Alam.
\newblock A multi-sensor information fusion approach for efficient 3d
  reconstruction in smart phone.
\newblock In {\em International Conference on Image Processing, Computer
  Vision, and Pattern Recognition}, 2015.

\bibitem{mcleod2017analysis}
A~Jonathan McLeod, Dante~PI Capaldi, John~SH Baxter, Grace Parraga, Xiongbiao
  Luo, and Terry~M Peters.
\newblock Analysis of periodicity in video sequences through dynamic linear
  modeling.
\newblock In {\em International Conference on Medical Image Computing and
  Computer-Assisted Intervention (MICCAI)}, pages 386--393. Springer, 2017.

\bibitem{nogueira2020analysis}
Mariana Nogueira, Mathieu De~Craene, Sergio Sanchez-Martinez, Devyani
  Chowdhury, Bart Bijnens, and Gemma Piella.
\newblock Analysis of nonstandardized stress echocardiography sequences using
  multiview dimensionality reduction.
\newblock {\em Medical Image Analysis}, 60:101594, 2020.

\bibitem{owayjan2012design}
Michel Owayjan, Ahmad Kashour, Nancy Al~Haddad, Mohamad Fadel, and Ghinwa
  Al~Souki.
\newblock The design and development of a lie detection system using facial
  micro-expressions.
\newblock In {\em International Conference on Advances in Computational Tools
  for Engineering Applications (ACTEA)}, pages 33--38. IEEE, 2012.

\bibitem{papadias2000globally}
Constantinos~B Papadias.
\newblock Globally convergent blind source separation based on a multiuser
  kurtosis maximization criterion.
\newblock {\em IEEE Transactions on Signal Processing}, 48(12):3508--3519,
  2000.

\bibitem{phung2002novel}
Son~Lam Phung, Abdesselam Bouzerdoum, and Douglas Chai.
\newblock A novel skin color model in ycbcr color space and its application to
  human face detection.
\newblock In {\em International Conference on Image Processing (ICIP)},
  volume~1, pages I--289. IEEE, 2002.

\bibitem{poh2011advancements}
Ming-Zher Poh, Daniel~J McDuff, and Rosalind~W Picard.
\newblock Advancements in noncontact, multiparameter physiological measurements
  using a webcam.
\newblock {\em IEEE Transactions on Biomedical Engineering}, 58(1):7--11, 2011.

\bibitem{qiu2018evm}
Ying Qiu, Yang Liu, Juan Arteaga-Falconi, Haiwei Dong, and Abdulmotaleb
  El~Saddik.
\newblock {EVM-CNN}: Real-time contactless heart rate estimation from facial
  video.
\newblock {\em IEEE Transactions on Multimedia}, 21(7):1778--1787, 2019.

\bibitem{rapczynski2019effects}
Michal Rapczynski, Philipp Werner, and Ayoub Al-Hamadi.
\newblock Effects of video encoding on camera based heart rate estimation.
\newblock {\em IEEE Transactions on Biomedical Engineering}, 66(12):3360--3370,
  2019.

\bibitem{rapczynski2018region}
Michal Rapczynski, Philipp Werner, Frerk Saxen, and Ayoub Al-Hamadi.
\newblock How the region of interest impacts contact free heart rate estimation
  algorithms.
\newblock In {\em International Conference on Image Processing (ICIP)}, pages
  2027--2031. IEEE, 2018.

\bibitem{reichert1992automatic}
Juergen Reichert.
\newblock Automatic classification of communication signals using higher order
  statistics.
\newblock In {\em International Conference on Acoustics, Speech, and Signal
  Processing}, volume~5, pages 221--224. IEEE, 1992.

\bibitem{rodriguez2018video}
Angel~Melchor Rodr{\'\i}guez and J~Ramos-Castro.
\newblock Video pulse rate variability analysis in stationary and motion
  conditions.
\newblock {\em Biomedical engineering online}, 17(1):11, 2018.

\bibitem{ross2006handbook}
Arun~A Ross, Karthik Nandakumar, and Anil Jain.
\newblock {\em Handbook of multibiometrics}, volume~6.
\newblock Springer Science \& Business Media, 2006.

\bibitem{salahuddin2007ultra}
Lizawati Salahuddin, Jaegeol Cho, Myeong~Gi Jeong, and Desok Kim.
\newblock Ultra short term analysis of heart rate variability for monitoring
  mental stress in mobile settings.
\newblock In {\em International Conference of the IEEE Engineering in Medicine
  and Biology Society (EMBC)}, pages 4656--4659. IEEE, 2007.

\bibitem{saragih2011deformable}
Jason~M Saragih, Simon Lucey, and Jeffrey~F Cohn.
\newblock Deformable model fitting by regularized landmark mean-shift.
\newblock {\em International Journal of Computer Vision}, 91(2):200--215, 2011.

\bibitem{Broj1}
Sanjana Sinha, Brojeshwar Bhowmick, Kingshuk Chakravarty, Aniruddha Sinha, and
  Abhijit Das.
\newblock Accurate upper body rehabilitation system using kinect.
\newblock In {\em International Conference of the IEEE Engineering in Medicine
  and Biology Society}, pages 4605--4609, 2016.

\bibitem{soleymani2011multimodal}
Mohammad Soleymani, Jeroen Lichtenauer, Thierry Pun, and Maja Pantic.
\newblock A multimodal database for affect recognition and implicit tagging.
\newblock {\em IEEE Transactions on Affective Computing}, 3(1):42--55, 2012.

\bibitem{tang2019removing}
Emily J Lam~Po Tang, Amir HajiRassouliha, Martyn~P Nash, Andrew~J Taberner,
  Poul~MF Nielsen, and Yusuf~O Cakmak.
\newblock Removing drift from carotid arterial pulse waveforms: A comparison of
  motion correction and high-pass filtering.
\newblock In {\em International Conference on Medical Image Computing and
  Computer-Assisted Intervention (MICCAI)}, pages 111--119. Springer, 2019.

\bibitem{tarvainen2002advanced}
Mika~P Tarvainen, Perttu~O Ranta-Aho, Pasi~A Karjalainen, et~al.
\newblock An advanced detrending method with application to {HRV} analysis.
\newblock {\em IEEE Transactions on Biomedical Engineering}, 49(2):172--175,
  2002.

\bibitem{tulyakov2016self}
Sergey Tulyakov, Xavier Alameda-Pineda, Elisa Ricci, Lijun Yin, Jeffrey~F Cohn,
  and Nicu Sebe.
\newblock Self-adaptive matrix completion for heart rate estimation from face
  videos under realistic conditions.
\newblock In {\em IEEE Conference on Computer Vision and Pattern Recognition
  (CVPR)}, pages 2396--2404, 2016.

\bibitem{verkruysse2008remote}
Wim Verkruysse, Lars~O Svaasand, and J~Stuart Nelson.
\newblock Remote plethysmographic imaging using ambient light.
\newblock {\em Optics express}, 16(26):21434--21445, 2008.

\bibitem{viola2001rapid}
Paul Viola and Michael Jones.
\newblock Rapid object detection using a boosted cascade of simple features.
\newblock In {\em IEEE Conference on Computer Vision and Pattern Recognition
  (CVPR)}, pages 511--518. IEEE, 2001.

\bibitem{wang2014facial}
Nannan Wang, Xinbo Gao, Dacheng Tao, and Xuelong Li.
\newblock Facial feature point detection: A comprehensive survey.
\newblock {\em arXiv preprint arXiv:1410.1037}, 2014.

\bibitem{wang2005automatic}
Peng Wang, Matthew~B Green, Qiang Ji, and James Wayman.
\newblock Automatic eye detection and its validation.
\newblock In {\em IEEE Conference on Computer Vision and Pattern
  Recognition-Workshops (CVPRW)}, pages 164--164. IEEE, 2005.

\bibitem{wang2016algorithmic}
Wenjin Wang, Albertus~C den Brinker, Sander Stuijk, and Gerard de~Haan.
\newblock Algorithmic principles of remote ppg.
\newblock {\em IEEE Transactions on Biomedical Engineering}, 64(7):1479--1491,
  2016.

\bibitem{Wu12Eulerian}
Hao-Yu Wu, Michael Rubinstein, Eugene Shih, John Guttag, Fr\'{e}do Durand, and
  William~T. Freeman.
\newblock Eulerian video magnification for revealing subtle changes in the
  world.
\newblock {\em ACM Transactions on Graphics}, 31(4), 2012.

\bibitem{yu2006method}
Chenggang Yu, Zhenqiu Liu, Thomas McKenna, Andrew~T Reisner, and Jaques
  Reifman.
\newblock A method for automatic identification of reliable heart rates
  calculated from {ECG} and {PPG} waveforms.
\newblock {\em Journal of the American Medical Informatics Association},
  13(3):309--320, 2006.

\bibitem{zhang2015troika}
Zhilin Zhang, Zhouyue Pi, and Benyuan Liu.
\newblock \mbox{TROIKA}: A general framework for heart rate monitoring using
  wrist-type photoplethysmographic signals during intensive physical exercise.
\newblock {\em IEEE Transactions on biomedical engineering}, 62(2):522--531,
  2015.

\end{thebibliography}
\end{document}